# Moral Lineage Tracing

Florian Jug[1,*], Evgeny Levinkov[2,*], Corinna Blasse[1], Eugene W. Myers[1], Bjoern Andres[2,†]

**Abstract**

Lineage tracing, the tracking of living cells as they move and divide, is a central problem in biological image analysis. Solutions, called lineage forests, are key to understanding how the structure of multicellular organisms emerges. We propose an integer linear program (ILP) whose feasible solutions define, for every image in a sequence, a decomposition into cells (segmentation) and, across images, a lineage forest of cells (tracing). In this ILP, path-cut inequalities enforce the morality of lineages, i.e., the constraint that cells do not merge. To find feasible solutions of this NP-hard problem, with certified bounds to the global optimum, we define efficient separation procedures and apply these as part of a branch-and-cut algorithm. To show the effectiveness of this approach, we analyze feasible solutions for real microscopy data in terms of bounds and run-time, and by their weighted edit distance to lineage forests traced by humans.

## 1 Introduction

Phenomenal progress in microscopy allows biologists to image large numbers of living cells as they move and divide [30, 45]. Such observations are essential in developmental biology for studying embryogenesis and tissue formation [31, 33, 37]. Consequently, the tracing of cells and their lineages in sequences of images has become a central problem in biological image analysis [4, 5, 29].

The lineage tracing problem consists of two sub-problems. The first sub-problem is to identify the cells in every individual image. The second sub-problem is to connect every cell identified in an image to the same cell and descendant cells identified in subsequent images. A joint solution of both sub-problems is a set of pairwise disjoint lineage trees (depicted in Fig. 1, in red and green) whose nodes are cells.

The first sub-problem is an image decomposition problem: If every pixel shows a part of a cell and no pixel shows a background, the objective is to decompose the pixel grid graph of the image into precisely one component per cell. If pixels potentially show background, the objective is to jointly select and decompose a subgraph of the pixel grid graph such that there is precisely one component for each cell and no component for the background.

The second sub-problem is a cell tracking problem: The objective is to connect every cell detected in one image to the same cell and descendant cells identified in subsequent images. A joint solution of both sub-problems is constrained by prior knowledge. In particular, every cell has at most one direct progenitor cell, i.e., cells do not merge. Moreover, no cell splits into more than two cells at once. Yet, a cell can appear without a direct progenitor cell when entering the field of view, and a cell can disappear when dying or leaving the field of view. Finally, it can appear as if a cell was dividing into more than two cells at once if the temporal

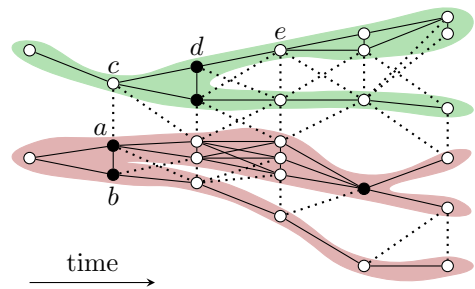

**Figure 1** Given a sequence of images, taken at consecutive points in time, and given a decomposition of each image into *cell fragments* (depicted above as nodes), the objective of lineage tracing is to join fragments of the same cell within and across images, e.g. $\{a,b\}$ and $\{c,d\}$, and to join fragments of descendant cells across images, e.g. $\{d,e\}$. Joins (cuts) are depicted as solid (dotted) lines. Fragments of dividing cells are depicted as black nodes.

resolution is too low to separate consecutive divisions.

It is understood that errors in the image decomposition make it harder to reconstruct the true lineage forest. Attempts at reconstructing the lineage forest can help to avoid such errors. Thus, we state a joint optimization problem whose feasible solutions define, for every image, a decomposition (segmentation) into cells and, across images, a lineage forest. Unlike in prior work, we do not constrain the set of decompositions, except by contracting pixels to superpixels.

## 2 Related Work

The image decomposition problem has been tackled by various abstractions in the form of optimization problems including the minimum cost spanning forest problem, i.e., agglomerative clustering [1], balanced cut problems, i.e., spectral clustering [10, 43], and the minimum cost multicut problem, i.e., correlation clustering [34]. We build on its formulation as a minimum cost multicut problem, an optimization problem studied in [16, 18] which is NP-hard [12, 17] and has been used for image segmentation

---

[1] MPI of Molecular Cell Biology and Genetics, Dresden
[2] MPI for Informatics, Saarbrücken
* Contributed equally
† Correspondence: andres@mpi-inf.mpg.de

in [3, 7, 8, 9, 11, 13, 14, 25, 26, 27, 32, 34, 51, 52].

The lineage forest reconstruction problem has been cast as an optimization problem in [21, 24, 28, 39, 41, 46, 47, 48, 49]. If cells neither die nor enter or leave the field of view and if one drops the constraint that cells split into at most two direct descendant cells, the problem can be formulated as a minimum cost $k$ disjoint arborescence problem [42, Section 53.9], as shown in [48, 49]. Here, $k$ is the number of cells visible in the first image. This problem can be solved in strongly polynomial time [20]. With the additional constraint that cells split into at most two descendant cells, the problem becomes NP-hard, and so do generalizations [21, 24, 28, 39, 41, 46, 47] that model, *e.g.*, the (dis)appearance of cells.

One lineage tracing approach [28] copes with imperfect decompositions by over-segmenting individual images. This guarantees that every cell is represented by at least one component and that every component represents at most one cell. Advantageous there is the fact that the true lineage forest is represented by at least one set of disjoint arborescences. Disadvantageous is the loss of robustness: For the true decomposition, every component belongs to precisely one arborescence and thus, every error in the set of disjoint arborescences implies at least a second error. This renders solutions robust to perturbations of the objective function. For an over-segmentation, this property is lost. Another disadvantage is the fact that the number of progenitor cells is not determined by the number of components of the first image. Over-estimates result in excessive arborescences that typically conflict with correct ones. Under-estimates result in a loss of lineage trees. As in [28], we consider an over-decomposition of each image into cell fragments. In contrast to [28] where each node of a lineage forest is a single representative fragment, each node in the lineage forests we consider is a clusters of fragments. This idea of clustering instead of selecting has been used in [44] to track multiple people in a video sequence. The optimization problem defined in [44] is a hybrid of a minimum cost multicut problem and a disjoint path problem. The optimization problem we propose here is a hybrid of a minimum cost multicut problem and a disjoint arborescence problem.

Two techniques have been proposed to deal with over and under-decomposition simultaneously: The first [41] is to allow single image components to represent multiple cells and thus be part of multiple lineages. This relaxation of the disjointness constraint of the arborescence problem introduces additional feasible solutions that can represent the true lineage forest even in the presence of under-decomposition. The same idea is used in [47] for the reconstruction of curvilinear structures and in [50] for the tracking of objects in containers. The second technique [21, 24, 40] considers a hierarchy of alternative decompositions and casts lineage tracing as an optimization problem whose feasible solutions select and connect components from the hierarchy. Constraints guarantee that selected components are mutually consistent and consistent with a set of disjoint lineages. As in [21, 24], the feasible solutions we propose define,

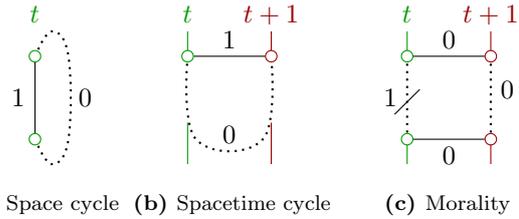

**(a)** Space cycle **(b)** Spacetime cycle **(c)** Morality

**Figure 2** Depicted above are examples (graphs and 01-labelings of edges) in which inequalities (1)-(3) are violated. **(a)** An inequality (1) is violated iff there exist $t \in \mathbb{N}$ and a cycle $Y$ in $G_t$ in which precisely one edge is labeled 1. **(b)** An inequality (2) is violated iff there exist $t \in \mathbb{N}$, an edge $\{v, w\} \in E_{t,t+1}$ labeled 1 and a path in $G_t^+$ connecting $v$ to $w$ in which all edges are labeled 0. **(c)** An inequality (3) is violated iff there exist $t \in T$ and nodes $v_t, w_t \in V_t$ and $v_{t+1}, w_{t+1} \in V_{t+1}$ connected by edges $\{v_t, v_{t+1}\}, \{w_t, w_{t+1}\} \in E_{t,t+1}$ labeled 0, such that $v_t$ and $w_t$ are separated by a cut in $G_t$ with all edges labeled 1 and $v_{t+1}$ and $w_{t+1}$ are connected by a path in $G_{t+1}$ with all edges labeled 0.

for every image, a decomposition into cells and, across images, a lineage forest. In contrast to prior work, we do not constrain the set of decompositions, except by contracting pixels to superpixels. We compare our experimental results to a state-of-the-art software system for lineage tracing [2].

Unlike in the work discussed above, feasible lineages and costs of feasible lineages can be defined recursively, as in particle filtering. Cf. [15] for a recent comprehensive comparison and [5] for a recent application to lineage tracing.

## 3 Optimization Problem

In this section, we cast lineage tracing as an optimization problem that we call *moral lineage tracing*. In Section 3.1, we define the set of feasible solutions. In Section 3.2, we define the objective function and optimization problem.

*3.1 Feasible Set*

In order to encode a combinatorial number of feasible lineage forests, we define a *hypothesis graph*. In a hypothesis graph, every node corresponds to one superpixel of one image in a sequence and is referred to as a *cell fragment* (Def. 1). In order to encode a single feasible lineage forest, we define a *lineage graph* (Def. 2). A lineage graph is a subgraph of the hypothesis graph that defines, within each image, a clustering of cell fragments into cells and, across images, a lineage forest of cells (Lemma 1). In order to state in the form of an ILP an optimization problem whose feasible solutions are lineage graphs, we identify the characteristic functions of lineage graphs with 01-labelings of the edges of the hypothesis graph that satisfy a system of linear inequalities (Lemma 2).

**Definition 1** A *hypothesis graph* is a node-labeled graph[1] $G = (V, E, \tau)$ in which every edge $\{v, w\} \in E$ holds $|\tau(v) - \tau(w)| \leq 1$. Every $v \in V$ is called a *(cell) fragment*, and $\tau(v)$ is called its *time index*.

The intuition is this: For any distinct fragments $v, w \in V$ with $\tau(v) = \tau(w)$, the presence of the edge $\{v, w\} \in E$ indicates the possibility that $v$ and $w$ are fragments of the same cell. For any fragments $v, w \in V$ with $\tau(w) = \tau(v) + 1$, the presence of the edge $\{v, w\} \in E$ indicates the possibility that $v$ and $w$ are fragments of the same cell, observed at successive points in time, as well as the possibility that $v$ is a fragment of a progenitor cell of the cell of $w$.

Next, we characterize those subgraphs of a hypothesis graph that we consider as feasible solutions. For clarity, we propose some notation: For every $t \in \mathbb{N}$, let $V_t := \tau^{-1}(t)$ the set of all fragments having the time index $t$. Let $G_t = (V_t, E_t)$ be the subgraph of $G$ induced by $V_t$. Let $E_{t,t+1} := \{\{v, w\} \in E | v \in V_t \land w \in V_{t+1}\}$ be the set of those edges of $G$ that connect a fragment $v$ having the time index $t$ to a fragment $w$ having the time index $t+1$. Let $G_t^+ = (V_t^+, E_t^+)$ be the subgraph of $G$ induced by $V_t^+ := V_t \cup V_{t+1}$. Finally, let $V_{\geq t} := \cup_{t'=t}^{\infty} V_{t'}$ be the fragments of at least time index $t$, and $E_{\geq t} := \cup_{t'=t}^{\infty}(E_{t'} \cup E_{t',t'+1})$ the set of all edges of $G$ between such fragments.

**Definition 2** For every hypothesis graph $G = (V, E, \tau)$, a set $C \subseteq E$ is called a *lineage cut* of $G$, and $(V, \bar{C})$ with $\bar{C} := E \setminus C$ is called a *lineage (sub)graph* of $G$, iff the following conditions hold:

1. For every $t \in \mathbb{N}$, the set $E_t \cap C$ is a multicut[2] of $G_t$

2. For every $t \in \mathbb{N}$ and every $\{v, w\} \in E_{t,t+1} \cap C$, $v$ and $w$ are not connected by any path in the graph $(V_t^+, E_t^+ \cap \bar{C})$

3. For every $t \in \mathbb{N}$, any $v_t, w_t \in V_t$ and any $v_{t+1}, w_{t+1} \in V_{t+1}$ such that $\{v_t, v_{t+1}\} \in E \cap \bar{C}$ and $\{w_t, w_{t+1}\} \in E \cap \bar{C}$, and for any path in $(V, E_{t+1} \cap \bar{C})$ from $v_{t+1}$ to $w_{t+1}$, there exists a path in $(V, E_t \cap \bar{C})$ from $v_t$ to $w_t$.

If these conditions are satisfied then, for every $t \in \mathbb{N}$ and every non-empty, maximal connected subgraph $(V_t', E_t')$ of $(V_t, E_t \cap \bar{C})$, its node set $V_t'$ is called a *cell* at time index $t$.

A lineage cut and lineage subgraph are called *binary* iff, in addition to Conditions 1–3, it holds:

4. For every $t \in \mathbb{N}$, every cell $V_t' \subseteq V_t$ is connected in the lineage subgraph to at most two distinct cells at $t+1$.

An intuition for Conditions 1–3 is offered by Lemma 1 and the proof.

**Lemma 1** For every $t \in \mathbb{N}$, a lineage graph well-defines a decomposition of $G_t$ whose components are the cells at time index $t$. Across time, a lineage graph well-defines a (lineage) forest of cells.

---
[1] All graphs are assumed to be finite, simple and undirected. A node labeling of a graph $(V, E)$ is a map $\tau : V \to \mathbb{N}$.
[2] A multicut of $G_t = (V_t, E_t)$ is a subset $M \subseteq E_t$ of edges such that, for every cycle $Y$ in $G_t$: $|M \cap Y| \neq 1$ [16].

PROOF Condition 1 guarantees that every subgraph defining a cell is node-induced, i.e., for every $t \in \mathbb{N}$ and every $\{v, w\} \in E_t$: $\{v, w\} \in \bar{C}$ iff $v$ and $w$ are fragments of the same cell. Condition 2 guarantees, for every $t \in \mathbb{N}$, every cell $V_t'$ at time index $t$, and every cell $V_{t+1}'$ at time index $t+1$ that either all edges of $G$ between $V_t'$ and $V_{t+1}'$ are in $\bar{C}$, or none. Condition 3 guarantees, for every $t \in \mathbb{N}$ and every distinct cells $V_t', V_t''$ at time index $t$ that these are not connected in $(V, E_t^+ \cap \bar{C})$ to the same cell at time index $t+1$. This guarentees, by induction, that $V_t', V_t''$ are not connected by any path in the graph $(V_{\geq t}, E_{\geq t} \cap \bar{C})$. This guarantees that distinct cells never merge. □

**Lemma 2** For every hypothesis graph $G = (V, E, \tau)$ and every $x \in \{0, 1\}^E$, the set $x^{-1}(1)$ of edges labeled 1 is a lineage cut of $G$ iff $x$ satisfies the linear inequalities (1)–(3) stated below. It is sufficient in (1) to consider only chordless cycles. Moreover, the lineage cut is binary iff, in addition, $x$ satisfies the linear inequality (4).

$\forall t \in \mathbb{N} \ \forall Y \in \text{cycles}(G_t) \ \forall e \in Y:$
$$x_e \leq \sum_{e' \in Y \setminus \{e\}} x_{e'} \tag{1}$$

$\forall t \in \mathbb{N} \ \forall \{v, w\} \in E_{t,t+1} \ \forall P \in vw\text{-paths}(G_t^+):$
$$x_{vw} \leq \sum_{e \in P} x_e \tag{2}$$

$\forall t \in \mathbb{N} \ \forall \{v_t, v_{t+1}\}, \{w_t, w_{t+1}\} \in E_{t,t+1}$
$\forall T \in v_t w_t\text{-cuts}(G_t) \ \forall P \in v_{t+1} w_{t+1}\text{-paths}(G_{t+1}):$
$$1 - \sum_{e \in T}(1 - x_e) \leq x_{v_t v_{t+1}} + x_{w_t w_{t+1}} + \sum_{e \in P} x_e \tag{3}$$

$\forall t \in \mathbb{N} \ \forall v \in V_t \ \forall w_1, w_2, w_3 \in V_{t+1}$
$\forall P_1 \in vw_1\text{-paths}(G_t^+) \ \forall P_2 \in vw_2\text{-paths}(G_t^+)$
$\forall P_3 \in vw_3\text{-paths}(G_t^+) \ \forall C_{12} \in w_1 w_2\text{-cuts}(G_{t+1})$
$\forall C_{23} \in w_2 w_3\text{-cuts}(G_{t+1}) \ \forall C_{13} \in w_1 w_3\text{-cuts}(G_{t+1}):$
$$1 - \sum_{e \in C_{12} \cup C_{23} \cup C_{13}}(1 - x_e) \leq \sum_{e \in P_1 \cup P_2 \cup P_3} x_e \tag{4}$$

A proof of Lemma 2 is given in the Appendix A. Complementary to the proof, a discussion of (dis)-connectedness w.r.t. a multicut can be found in [6].

Here, the set of all $x \in \{0, 1\}^E$ that satisfy (1)–(4) is denoted by $X_G$. Examples of violated inequalities are depicted in Fig. 2. Note that the path-cut inequalities (3) guarantee that any fragmets of the same cell at time $t+1$ cannot be joined with fragments of distinct cells at time $t$, *i.e.*, morality.

*3.2 Objective Function*

**Definition 3** A *priced hypothesis graph* is a tuple $(V, E, \tau, c, c^+, c^-)$ with $(V, E, \tau)$ a hypothesis graph, $c : E \to \mathbb{R}$ and $c^+, c^- : V \to \mathbb{R}_0^+$. For every $e \in E$, $c_e$ is called the *cut cost* of $e$. For every $v \in V$, $c_v^+$ and $c_v^-$ are called the *appearance* and *disappearance cost* of $v$, respectively.

The optimization problem we propose is defined below in the form of an ILP w.r.t. a priced hypothesis graph $G = (V, E, \tau, c, c^+, c^-)$. This ILP has the following properties: Every feasible solutions defines a lineage subgraph of $G$. For every $\{v, w\} = e \in E$, the objective function assigns the cost (or reward) $c_e$ to all lineage graphs in which the cell fragments $v$ and $w$ belong to distinct cells. For every $t \in \mathbb{N}$ and every $v \in V_{t+1}$, the objective function assigns the (appearance) cost $c_v^+$ to all lineage graphs in which the fragment $v$ is not joined with any fragment in $V_t$. For every $t \in \mathbb{N}$ and every $v \in V_t$, the objective function assigns the (disappearance) cost $c_v^-$ to all lineage graphs in which the fragment $v$ is not joined with any fragment in $V_{t+1}$.

**Definition 4** For any priced hypothesis graph $G = (V, E, \tau, c, c^+, c^-)$, the instance of the *moral lineage tracing problem* w.r.t. $G$ is the ILP in $x \in \{0,1\}^E$ and $x^+, x^- \in \{0,1\}^V$ written below.

$$\min_{x, x^+, x^-} \sum_{e \in E} c_e x_e + \sum_{v \in V} c_v^+ x_v^+ + \sum_{v \in V} c_v^- x_v^- \quad (5)$$

$$\text{subject to} \quad x \in X_G \quad (6)$$

$$\forall t \in \mathbb{N} \ \forall v \in V_{t+1} \ \forall T \in V_t v\text{-cuts}(G_t^+):$$
$$1 - x_v^+ \leq \sum_{e \in T} (1 - x_e) \quad (7)$$

$$\forall t \in \mathbb{N} \ \forall v \in V_t \ \forall T \in vV_{t+1}\text{-cuts}(G_t^+):$$
$$1 - x_v^- \leq \sum_{e \in T} (1 - x_e) \quad (8)$$

If, in an inequality of (7), all edges in the cut $T$ are labeled 1, then $x_v^+ = 1$. Otherwise, for every feasible solution $x$ the same solution but with $x_v^+ := 0$ is not worse (as $0 \leq c_v^+$, by definition of $c^+$). Thus, a cost $c_v^+ \neq 0$ is payed iff fragment $v$ appears at time $t + 1$. The argument for (8) and disappearance is analogous.

## 4 Optimization Algorithm

### 4.1 Efficient Separation Procedures

Below, we define, for each class of inequalities, (1)–(4), (7) and (8), an efficient separation procedure (Tab. 1) that takes any $(x, x^+, x^-)$ as input. If any inequality is violated, it terminates and outputs at least one of these. If no inequality is violated, it terminates and outputs the empty set. We apply these procedures in a branch-and-cut algorithm described in the next section. We have also applied these procedures in the preparation of the experiments described in Section 5, to certify the well-definedness of lineages we traced manually.

To separate infeasible solutions by inequalities (1) for a given $t$, we label maximal subgraphs of $G_t$ connected by edges labeled 0. Then, for every $\{v, w\} = e \in E_t$ with $x_e = 1$ and with $v$ and $w$ being in the same subgraph, we search for a shortest $vw$-path $P$ in $G_t$ such that $x_P = 0$, using breadth-first-search (BFS). If the path is chordless, we output the inequality defined by the cycle $P \cup \{e\}$ and $e$.

To separate infeasible solutions by inequalities (2) for a given $t$, we label maximal subgraphs of $G_t^+$ connected by edges labeled 0. Then, for every $\{v, w\} = e \in E_{t, t+1}$ with $x_e = 1$ and with $v$ and $w$ being in the same subgraph, we search for a shortest $vw$-path $P$ in $G_t^+$ such that $x_P = 0$ using BFS. We output the inequality defined by the cycle $Y := P \cup \{e\}$ and $e \in Y$.

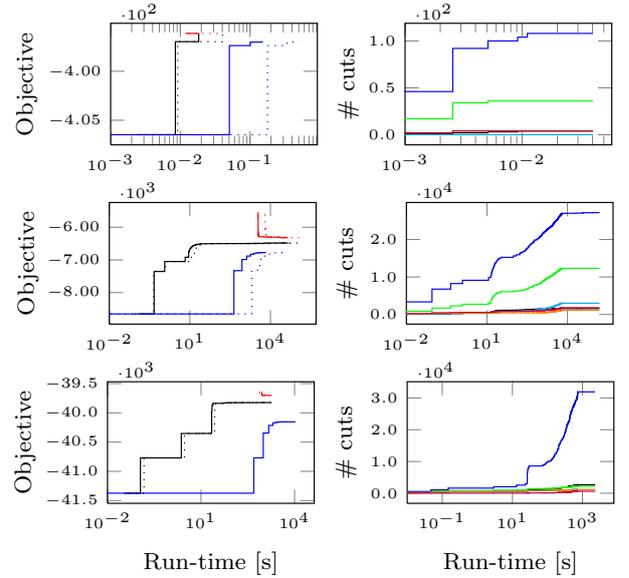

**Figure 3** Depicted above is the convergence of the branch-and-cut algorithm that solves the ILP and of the cutting plane algorithm that solves the canonical LP relaxation. Instances of the moral lineage tracing problem are, from top to bottom, *HeLa-small*, *HeLa-test* and *Flywing*. Graphs on the left show the objective values of intermediate integer feasible solutions (–) and lower bounds (–) found by the branch-and-cut algorithm, as well as the lower bound found by the cutting-plane algorithm (–). Dotted lines show the convergence with the bifurcation constraint (4). It can be seen that the LP relaxation is not tight for larger problems. Graphs on the right show numbers of cuts: morality (–), spacetime cycle (–), space cycle (–), appearance (–), disappearance (–), and bifurcation (–). It can be seen that violated morality constraints dominate.

To separate infeasible solutions by inequalities (3) for a given $t$, we label maximal subgraphs of $G_t$ connected by edges labeled 0. Then, for every pair $v, w \in V_t$ of nodes with different labels, we use BFS to search for (i) a shortest $vw$-path $P$ in $(V_{t+1}, E_{t,t+1} \cup E_{t+1})$ such that $x_P = 0$, and (ii) a $vw$-cut $T$ in $G_t$ such that $x_T = 1$. We output the inequality defined by $P$ and $T$.

To separate infeasible solutions by inequalities (4) for a given $t$, we label maximal subgraphs of $G_t$ and $G_{t+1}$, resp., connected by edges labeled 0. From every $v \in V_t$, we start a BFS in the subgraph of $G_t^+$ whose edges are labeled 0. If nodes $w_1, w_2, w_3 \in V_{t+1}$ of three distinct components of $G_{t+1}$ are reached, we output the inequality defined by the boundaries of these components and by paths from $v$ to $w_j$.

To separate infeasible solutions by inequalities (7) for a given $t$, we start a BFS from every $v \in V_{t+1}$ in the subgraph of $G_t^+$ whose edges are labeled 0. We either find a vertex $w \in V_t$ (no violation) or a cut $T \in V_t v\text{-cuts}(G_t^+)$ which separates $v$ from $V_t$. In the latter case, we output the inequality defined by the vertex $v$ and the cut $T$. The separation of infeasible solutions by inequalities (8) is analogous, in the opposite order of time indices.

| Constraint | ILP (branch-and-cut) | LP (cutting-plane) |
|---|---|---|
| Space | $O(|E_t|^2)$ | $O(|E_t|^2 \log |V_t|)$ |
| Spacetime | $O(|E_{t,t+1}||E_t^+|)$ | $O(|E_{t,t+1}||E_t^+| \log |V_t^+|)$ |
| Morality | $O(|V_t|^2|E_t^+|)$ | $O(|V_t|^2(m(|V_t|,|E_t|)$ $+|E_{t+1}|\log |V_{t+1}|))$ |
| Termination | $O(|V_t| + |E_t^+|)$ | $O(|V_t|m(|V_t^+|,|E_t^+|))$ |
| Birth | $O(|V_{t+1}| + |E_t^+|)$ | $O(|V_{t+1}|m(|V_t^+|,|E_t^+|))$ |
| Bifurcation | $O(|E|\log |V| + |V_t|$ $+|E_t^+|\log |V_t^+|)$ | $O(|V_t||V_{t+1}|^3(m(|V_{t+1}|,|E_{t+1}|)$ $+|E_t^+|\log |V_t^+|))$ |

**Table 1** Worst case time complexity of the separation procedures we implement for integral points (ILP) and fractional points (LP). Here, $m(|V|,|E|)$ denotes the worst case time complexity of a maximum $st$-flow algorithm for a graph $(V,E)$.

| Problem Instance | Variables | Fractional | Optimal |
|---|---|---|---|
| HeLa-small | 1839 | 5 | 1834 (100%) |
| HeLa-test | 41571 | 1180 | 40078 (99.2%) |
| Flywing | 29063 | 1174 | 27740 (99.5%) |

**Table 2** Analysis of solutions of the canonical LP relaxation of the moral lineage tracing problem.

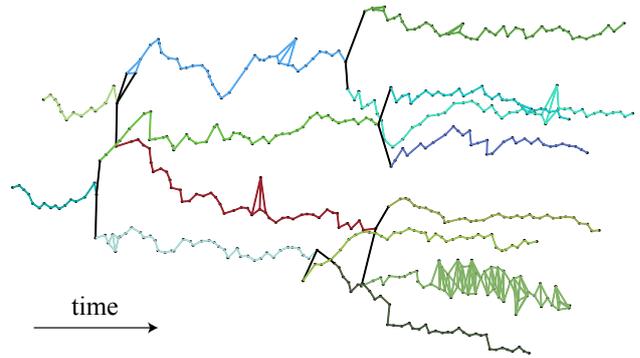

**Figure 4** Depicted above is the lineage forest $(V, \bar{C})$ reconstructed by solving an instance of the moral lineage tracing problem (Def. 4) defined w.r.t. the image sequence HeLa-small. Edges connecting a fragment of one cell to a fragment of a descendant cell (depicted in black) indicate cell divisions. Edges connecting fragments of the same cell are depicted in a color representing that cell. Note the two progenitor cells in the first image, visible here on the l.h.s..

### 4.2 Branch-and-Cut Algorithm for the ILP

In order to find feasible solutions of the moral lineage tracing problem (Def. 4), with certified bounds, we implement the separation procedures defined in the previous section in C++ and call these form the branch-and-cut algorithm of the ILP solver Gurobi [22] whenever an *integer* feasible solution is found. In order to tighten intermediate LP relaxations, we resort to the cuts implemented in Gurobi.

In all experiments we conduct, less than 1% of the total run-time is spent on the separation of infeasible solutions by inequalities (1)–(4), (7) and (8) together. Objective values, bounds and numbers of added inequalities are shown w.r.t. run-time, for three instances of the problem, in Fig. 3.

### 4.3 Cutting-Plane Algorithm for an LP Relaxation

In addition to the moral lineage tracing ILP and (integer) feasible solutions found by the branch-and-cut algorithm, we study the canonical LP relaxation and its (possibly fractional) solutions found by a cutting-plane algorithm. Results shown in Fig. 3 and Tab. 2 are discussed below:

It can be seen from Fig. 3 that the solution found by our cutting-plane algorithm for the LP (blue) converges slower than the lower bound found by our branch-and-cut algorithm for the ILP (black). This is simply because the separation of infeasible points by violated inequalities is more complex if the point can be fractional; see Tab. 1.

It can be seen from Fig. 3 and Tab. 2 that the LP relaxation is almost tight for the problem instance *HeLa-small* and less tight for the larger problem instances. This is expected, as *HeLa-small* is dominated by the disjoint arborescence sub-problem, which is in PTIME, while the larger problems are dominated by the minimum cost multicut sub-problem, which is NP-hard. The solution of the LP is not half-integral. Yet, it encourages future work on rounding procedures.

## 5 Application to Microscopy Data

In order to examine the effectiveness of moral lineage tracing (MLT) and the proposed branch-and-cut algorithm, we define three instances of the problem w.r.t. two biomedical data sets, *N2DL HeLa* and *Flywing Epithelium*.

### 5.1 N2DL HeLa Data

This microscopy data consists of three sequences of images which show HeLa cells that move and divide as bright objects in front of a dark background, *c.f.* Fig. 6(a). Two sequences are publicly available and one sequence is undisclosed for an annual competition [36]. Here, we use the two public sequences, one for learning a cost function, the other (*HeLa-test*) for experiments. To obtain, in addition, a shorter sequence of smaller images, we crop from *HeLa-test* a sub-problem (*HeLa-small*). For both sequences, we construct a priced hypothesis graph as shown in Fig. 5(a) and described below. The hypothesis graph for *HeLa-test* consists of 10882 nodes and 19807 edges. The hypothesis graph for *HeLa-small* consists of 512 nodes and 812 edges.

**Optimization.** The convergence of the branch-and-cut algorithm for the instances *HeLa-small* and *HeLa-test* of the MLT problem is shown in the first two rows of Fig. 3. It can be seen that the small problem is solved to optimality, while the full problem is solved with an optimality gap. Most separating cuts are morality constraints.

**Results.** A lineage forest for *HeLa-small* defined by the solution of the MLT problem is shown in Fig. 4. This lineage forest is in exact accordance with the ground truth. Corresponding decompositions of images are shown in Fig. 6. A lineage forest for *HeLa-test* defined by the feasible solution of the MLT problem is shown in Fig. 8(a). A comparison with ground truth provided in [36] is shown in Tab. 3 in terms of metrics SEG and TRA as defined in [36].

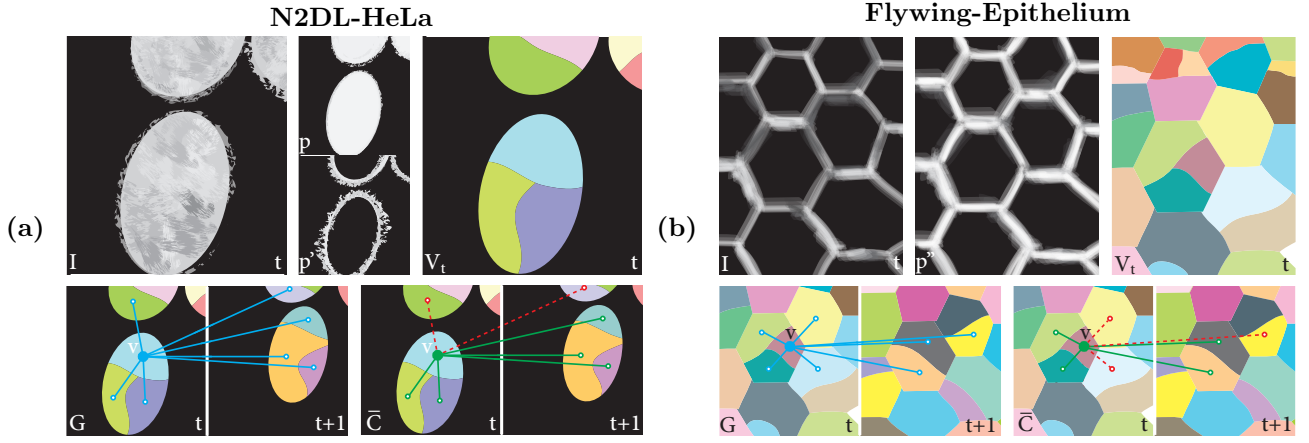

**Figure 5** Sketched above is the construction of an instance of the MLT for **(a)** the N2DL-HeLa Data and **(b)** the Flywing-Epithelium Data. For every time index $t$ and the respective image $I$ in the sequence, foreground probabilities $p$ of pixels showing part of a cell and probabilities $p'$ and $p''$ of pixels showing object boundaries are estimated and used to decompose the image into cell fragments $S_t$. A hypothesis graph (shown for only one fragment each) connects nearby cell fragments within images and across successive images.

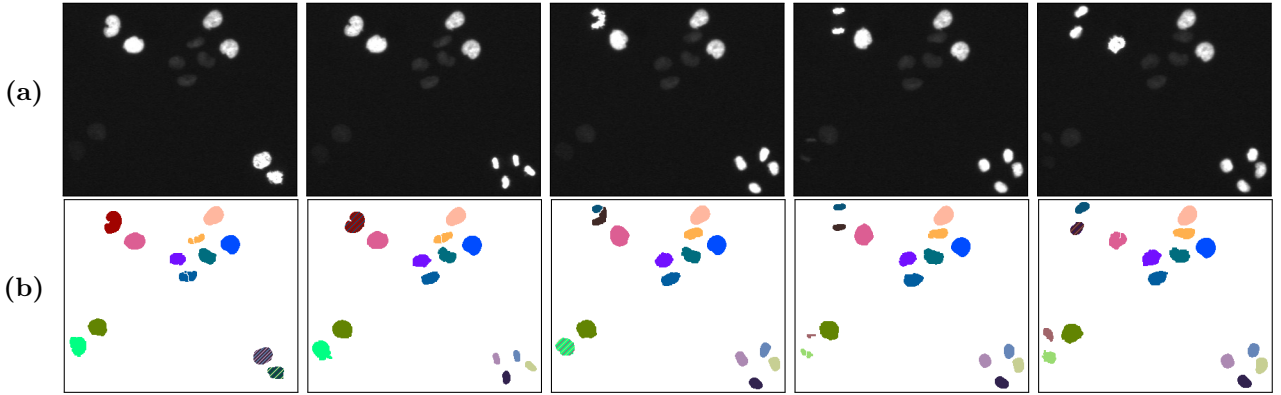

**Figure 6** Depicted above are **(a)** image crops of the full HeLa-test data set, and **(b)** decompositions of the images defined by a feasible solution of the moral lineage tracing problem (Def. 4). Diagonally striped cells indicate cell division.

**Technical details.** Before constructing a hypothesis graph we perform the following data pre-processing: We train a random forest to predict, for every pixel $r$, the probability $p_r$ of this pixel being foreground (part of a cell). For every time index $t$, we consider the set $S_t$ of pixels $r$ at time $t$ for which $p_r > 0.5$. A watershed search of the distance transform of $S_t$ decomposes the subgraph of the pixel grid graph induced by $S_t$ into cell fragments (superpixels) $V_t$. For every cell fragment $v \in V_t$, we compute its center of mass $r_v \in \mathbb{R}^2$ in the image plane. We train a second random forest to predict, for every pixel $r$, the probability $p'_r$ of this pixel showing a cell boundary, *i.e.* the interface between a cell and the background or the interface between two cells that touch.

We then construct a hypothesis graph $G = (V, E, \tau)$ as follows: For every time index $t$ and every pair of distinct cell fragments $v, w \in V_t$, we introduce the edge $\{v, w\} \in E_t$ iff $\|r_v - r_w\| < d_1$, for a maximum distance $d_1 \in \mathbb{R}^+$. For every time index $t$ and every pair $(v, w) \in V_t \times V_{t+1}$, we introduce the edge $\{v, w\} \in E_{t,t+1}$ iff $\|r_v - r_w\| < d_2$, for a maximum distance $d_2 \in \mathbb{R}^+$.

For every time index $t$ and every edge $e = \{v, w\} \in E_t$, we define the cost

$$c_e = -\text{logit} \max\{\|r_v - r_w\|/d_1, p'(r_v, r_w)\} \ . \quad (9)$$

Here $p'(r_v, r_w)$ denotes the maximum of $p'$ along the Bresenham line from $r_v$ to $r_w$. For every time index $t$ and every $e = \{v, w\} \in E_{t,t+1}$, we define the cost

$$c_e = -\text{logit} \|r_v - r_w\|/d_2 \ . \quad (10)$$

All (dis)appearance costs are constant, $c^+ = c^- = c_0 \in \mathbb{R}^+$.

### 5.2 Flywing Epithelium Data

This dataset contains images that show a developing fly wing epithelium, *c.f.* Fig. 7(a). Every pixel is a part of a cell and no pixels show background. The data is divided into a training sequence and a test sequence. We have collected ground truth for both sequences by manually joining watershed superpixels. The construction of a priced hypothesis graph from the raw test sequence is sketched in Fig. 5(b) and described in more detail below. It consists of 5026 nodes and 19011 edges.

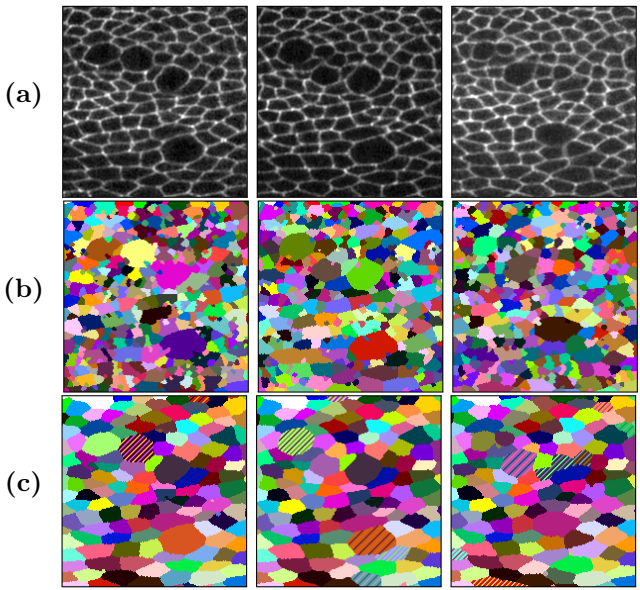

**Figure 7** Depicted above are **(a)** images of the fly wing test set, **(b)** decompositions of these images into cell fragments, and **(c)** decompositions of the images defined by a feasible solution of the moral lineage tracing problem (Def. 4). Diagonally striped cells divide before the next time point (image).

|  | Test Data | SEG | TRA |
| --- | --- | --- | --- |
| MLT (our) | public | 0.7811 | 0.9747 |
| KTH-SE [35] | undisclosed | 0.8932 | 0.9920 |
| HEID-GE [23] | undisclosed | 0.8155 | 0.9871 |
| LEID-NL [19] | undisclosed | 0.8180 | 0.9558 |
| HOUS-US [38] | undisclosed | 0.7701 | 0.9865 |
| NOTT-UK | undisclosed | 0.5778 | 0.7811 |
| IMCB-SG | undisclosed | 0.3317 | 0.9327 |

**Table 3** Quantified above is the distance from ground truth of decompositions (SEG) and lineage forests (TRA) obtained by MLT and contenders of the second ISBI Tracking Challenge [36]. Evaluation for [36] is performed on undisclosed test data. We evaluate MLT on test data published on the challenge website.

**Optimization.** The convergence of the branch-and-cut algorithm is shown in the third row of Fig. 3. It be seen from this figure that the problem is solved with an optimality gap determined by the lower bound.

**Results.** The lineage forest defined by the feasible solution of the problem is depicted in Fig. 8(b). Corresponding decompositions of images are depicted in Fig. 7(c). Decompositions and the lineage forests are compared in Tab. 4 to the ground truth in terms of the metrics SEG and TRA. It can be seen from this table that these results compare well to those found by the tracking system biologist use today [2].

**Technical details.** Data pre-processing consists of training a random forest classifier for detecting, for every pixel $r$, the probability $p''_r$ of showing a cell membrane. For every time index $t$, we decompose the image taken at time $t$ into cell fragments $V_t$ by first applying a watershed transform on the raw image sequence and then progressively joining adjacent superpixels iff both the average image intensity and the average membrane probability along their shared boundary are below respective thresholds. We maximize these thresholds w.r.t. the training data subject to the constraint that no false joins occur at this stage. This leads to $3.09 \pm 1.3$ fragments per cell. Also as pre-processing, we estimate dense optical flow $f$ for the image sequence and compute, for every cell fragment $v$, its center of mass $r_v \in \mathbb{R}^2$.

| Method | SEG | TRA |
| --- | --- | --- |
| MLT (our) | 0.9722 | 0.9813 |
| PA (on GT seg.) | 0.9327 | 0.9898 |
| PA (auto) | 0.7980 | 0.9206 |

**Table 4** Quantified above is the distance from ground truth of decompositions (SEG) and traced lineage forests (TRA) obtained by MLT and, alternatively, the *Packing Analyzer* [2].

We then construct a hypothesis graph $G = (V, E, \tau)$ as follows: For every time index $t$ and every pair of distinct cell fragments $v, w \in V_t$, we introduce the edge $\{v, w\} \in E_t$ iff $v$ and $w$ are adjacent components of the pixel grid graph of the image taken at time index $t$. For every time index $t$ and every pair $(v, w) \in V_t \times V_{t+1}$, we introduce the edge $\{v, w\} \in E_{t,t+1}$ iff $||r_v + f(r_v) - r_w||_2 \leq d$, for a maximum distance $d \in \mathbb{R}^+$.

For every time index $t$ and every edge $e = \{v, w\} \in E_t$, we define the cut-cost

$$c_e = -\text{logit} \sum_{r \in E(v,w)} p''(r)/|B(v,w)| \quad (11)$$

where $B(v, w)$ is the set of pixels in fragment $v$ adjacent to fragment $w$ and vice versa. For every time index $t$ and every edge $e = \{v, w\} \in E_{t,t+1}$, we define the cut-cost $c_e = c_0 + c_1 m_e$ with $m_e$ the maximum of $p''$ along a geodesic between pixels $r_v$ and $r_w$, and with $c_0, c_1 \in \mathbb{R}$ estimated from training data by logistic regression. All (dis)appearance costs are constant, $c^+ = c^- = c_0 \in \mathbb{R}^+$.

## 6 Conclusion

Building on recent work in image decomposition and multi-target tracking, we have proposed a rigorous mathematical abstraction of lineage tracing, a central problem in biological image analysis. The optimization problem we propose, a hybrid of the well-known minimum cost multicut problem and the minimum cost $k$ disjoint arborescence problem, is a joint formulation of image decomposition and lineage forest reconstruction. Its feasible solutions define, for every image in a sequence of images, a decomposition into cells and, across images, a lineage forest of cells. Unlike previous formulations, it does not constrain the set of decompositions. We have studied three instances of this problem defined by two biologically relevant microscopy data sets. For all instances, we have obtained feasible solutions with certified optimality gap. One instance has been solved to global optimality, yielding a solution in exact accordance with decompositions and ground truth lineages.

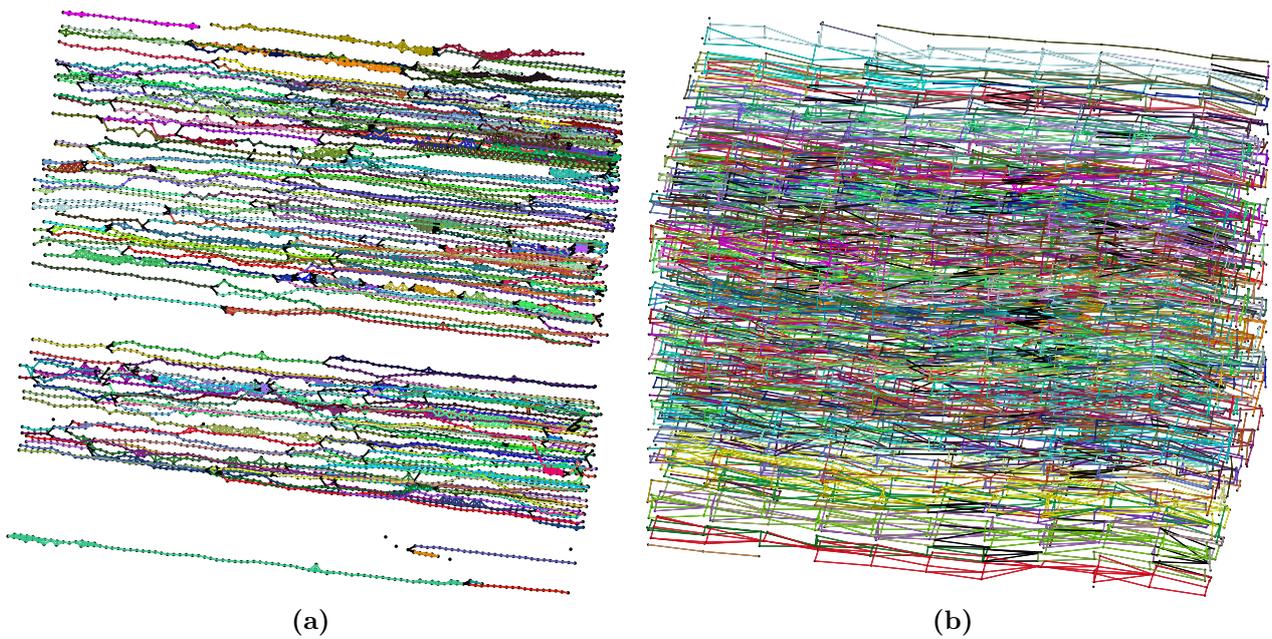

(a)        (b)

**Figure 8** 3D rendered lineage forests for **(a)** the full HeLa-test data set, and **(b)** the complete fly wing data set, as obtained by solving the moral lineage tracing problem. For better visibility, only the traced moral lineages are shown while all cut edges are hidden. Time progresses from left to right.

*Acknowledgements*

We thank the lab of Suzanne Eaton (MPI-CBG) for providing the flywing data. This work was partially supported by the German Federal Ministry of Research and Education (BMBF) under the code 031A099.

**References**

[1] R. Adams and L. Bischof. Seeded region growing. *TPAMI*, 16(6):641–647, June 1994.

[2] B. Aigouy, R. Farhadifar, D. B. Staple, A. Sagner, J.-C. Röper, F. Jülicher, and S. Eaton. Cell flow reorients the axis of planar polarity in the wing epithelium of Drosophila. *Cell*, 142(5):773–786, Sept. 2010.

[3] A. Alush and J. Goldberger. Ensemble segmentation using efficient integer linear programming. *TPAMI*, 34(10):1966–1977, 2012.

[4] F. Amat and P. J. Keller. Towards comprehensive cell lineage reconstructions in complex organisms using light-sheet microscopy. *Dev. Growth Differ.*, 55(4):563–578, May 2013.

[5] F. Amat, W. Lemon, D. P. Mossing, K. McDole, Y. Wan, K. Branson, E. W. Myers, and P. J. Keller. Fast, accurate reconstruction of cell lineages from large-scale fluorescence microscopy data. *Nature Methods*, 11(9):951–958, Sept. 2014.

[6] B. Andres. Lifting of multicuts. *CoRR*, abs/1503.03791, 2015. http://arxiv.org/abs/1503.03791.

[7] B. Andres, J. H. Kappes, T. Beier, U. Köthe, and F. A. Hamprecht. Probabilistic image segmentation with closedness constraints. In *ICCV*, 2011.

[8] B. Andres, T. Kröger, K. L. Briggman, W. Denk, N. Korogod, G. Knott, U. Köthe, and F. A. Hamprecht. Globally optimal closed-surface segmentation for connectomics. In *ECCV*, 2012.

[9] B. Andres, J. Yarkony, B. S. Manjunath, S. Kirchhoff, E. Turetken, C. C. Fowlkes, and H. Pfister. Segmenting planar superpixel adjacency graphs w.r.t. non-planar superpixel affinity graphs. In *EMMCVPR*, 2013.

[10] P. Arbeláez, M. Maire, C. Fowlkes, and J. Malik. Contour detection and hierarchical image segmentation. *TPAMI*, 33(5):898–916, 2011.

[11] S. Bagon and M. Galun. Large scale correlation clustering optimization. *CoRR*, abs/1112.2903, 2011.

[12] N. Bansal, A. Blum, and S. Chawla. Correlation clustering. *Machine Learning*, 56(1–3):89–113, 2004.

[13] T. Beier, F. A. Hamprecht, and J. H. Kappes. Fusion moves for correlation clustering. In *CVPR*, 2015.

[14] T. Beier, T. Kröger, J. H. Kappes, U. Köthe, and F. A. Hamprecht. Cut, Glue & Cut: A fast, approximate solver for multicut partitioning. In *CVPR*, 2014.

[15] N. Chenouard, I. Smal, F. de Chaumont, M. Maška, I. F. Sbalzarini, Y. Gong, J. Cardinale, C. Carthel, S. Coraluppi, M. Winter, A. R. Cohen, W. J.

Godinez, K. Rohr, Y. Kalaidzidis, L. Liang, J. Duncan, H. Shen, Y. Xu, K. E. G. Magnusson, J. Jaldén, H. M. Blau, P. Paul-Gilloteaux, P. Roudot, C. Kervrann, F. Waharte, J.-Y. Tinevez, S. L. Shorte, J. Willemse, K. Celler, G. P. van Wezel, H.-W. Dan, Y.-S. Tsai, C. Ortiz-de Solorzano, J.-C. Olivo-Marin, and E. Meijering. Objective comparison of particle tracking methods. *Nature Methods*, 11(3):281–289, Mar. 2014.

[16] S. Chopra and M. Rao. The partition problem. *Mathematical Programming*, 59(1–3):87–115, 1993.

[17] E. D. Demaine, D. Emanuel, A. Fiat, and N. Immorlica. Correlation clustering in general weighted graphs. *Theoretical Computer Science*, 361(2–3):172–187, 2006.

[18] M. M. Deza and M. Laurent. *Geometry of Cuts and Metrics*. Springer, 1997.

[19] O. Dzyubachyk, W. A. van Cappellen, J. Essers, W. J. Niessen, and E. Meijering. Advanced level-set-based cell tracking in time-lapse fluorescence microscopy. *IEEE Transactions on Medical Imaging*, 29(3):852–867, March 2010.

[20] J. Edmonds. Some well-solved problems in combinatorial optimization. In B. Roy, editor, *Combinatorial Programming: Methods and Applications*, volume 19 of *NATO Advanced Study Institutes Series*, pages 285–301. Springer, 1975.

[21] J. Funke, B. Anders, F. A. Hamprecht, A. Cardona, and M. Cook. Efficient automatic 3D-reconstruction of branching neurons from EM data. In *CVPR '12: Proceedings of the 2012 IEEE Conference on Computer Vision and Pattern Recognition (CVPR)*. IEEE Computer Society, June 2012.

[22] I. Gurobi Optimization. Gurobi optimizer reference manual, 2015.

[23] N. Harder, R. Batra, N. Diessl, S. Gogolin, R. Eils, F. Westermann, R. König, and K. Rohr. Large-scale tracking and classification for automatic analysis of cell migration and proliferation, and experimental optimization of high-throughput screens of neuroblastoma cells. *Cytometry Part A*, 87(6):524–540, June 2015.

[24] F. Jug, T. Pietzsch, D. Kainmüller, J. Funke, M. Kaiser, E. van Nimwegen, C. Rother, and G. Myers. Optimal Joint Segmentation and Tracking of Escherichia Coli in the Mother Machine. In *Bayesian and grAphical Models for Biomedical Imaging*, pages 25–36. Springer International Publishing, Cham, 2014.

[25] J. H. Kappes, M. Speth, B. Andres, G. Reinelt, and C. Schnörr. Globally optimal image partitioning by multicuts. In *EMMCVPR*, 2011.

[26] J. H. Kappes, M. Speth, G. Reinelt, and C. Schnörr. Higher-order segmentation via multicuts. *CoRR*, abs/1305.6387, 2013.

[27] J. H. Kappes, P. Swoboda, B. Savchynskyy, T. Hazan, and C. Schnörr. Probabilistic correlation clustering and image partitioning using perturbed multicuts. In *SSVM*, 2015.

[28] B. X. Kausler, M. Schiegg, B. Andres, M. Lindner, U. Koethe, H. Leitte, J. Wittbrodt, L. Hufnagel, and F. A. Hamprecht. A discrete chain graph model for 3d+t cell tracking with high mis-detection robustness. In *ECCV'12: Proceedings of the 12th European conference on Computer Vision*. Springer-Verlag, Oct. 2012.

[29] P. J. Keller. Imaging morphogenesis: technological advances and biological insights. *Science (New York, N.Y.)*, 340(6137):1234168, June 2013.

[30] P. J. Keller, A. D. Schmidt, A. Santella, K. Khairy, Z. Bao, J. Wittbrodt, and E. H. K. Stelzer. Fast, high-contrast imaging of animal development with scanned light sheet-based structured-illumination microscopy. *Nature Methods*, 7(8):637–642, Aug. 2010.

[31] P. J. Keller, A. D. Schmidt, J. Wittbrodt, and E. H. K. Stelzer. Reconstruction of zebrafish early embryonic development by scanned light sheet microscopy. *Science*, 322(5904):1065–1069, Nov. 2008.

[32] M. Keuper, E. Levinkov, N. Bonneel, G. Lavoué, T. Brox, and B. Andres. Efficient decomposition of image and mesh graphs by lifted multicuts. In *ICCV*, 2015.

[33] K. Khairy and P. J. Keller. Reconstructing embryonic development. *Genesis*, 49(7):488–513, July 2011.

[34] S. Kim, C. Yoo, S. Nowozin, and P. Kohli. Image segmentation using higher-order correlation clustering. *TPAMI*, 36:1761–1774, 2014.

[35] K. E. G. Magnusson, J. Jaldén, P. M. Gilbert, and H. M. Blau. Global linking of cell tracks using the Viterbi algorithm. *IEEE Trans. Med. Imag.*, 34(4):911–929, 2015.

[36] M. Maška, V. Ulman, D. Svoboda, P. Matula, P. Matula, C. Ederra, A. Urbiola, T. España, S. Venkatesan, D. M. W. Balak, P. Karas, T. Bolcková, M. Štreitová, C. Carthel, S. Coraluppi, N. Harder, K. Rohr, K. E. G. Magnusson, J. Jaldén, H. M. Blau, O. Dzyubachyk, P. Křížek, G. M. Hagen, D. Pastor-Escuredo, D. Jimenez-Carretero, M. J. Ledesma-Carbayo, A. Muñoz-Barrutia, E. Meijering, M. Kozubek, and C. Ortiz-de Solorzano. A benchmark for comparison of cell tracking algorithms. *Bioinformatics*, page btu080, Feb. 2014.

[37] S. G. Megason and S. E. Fraser. Imaging in systems biology. *Cell*, 130(5):784–795, Sept. 2007.

[38] A. Merouane, N. Rey-Villamizar, Y. Lu, I. Liadi, G. Romain, J. Lu, H. Singh, L. J. N. Cooper, N. Varadarajan, and B. Roysam. Automated profiling of individual cell–cell interactions from high-throughput time-lapse imaging microscopy in nanowell grids (TIMING). *Bioinformatics*, page btv355, June 2015.

[39] D. Padfield, J. Rittscher, and B. Roysam. Coupled minimum-cost flow cell tracking for high-throughput quantitative analysis. *Med Image Anal*, 15(4):650–668, Aug. 2011.

[40] M. Schiegg, P. Hanslovsky, C. Haubold, U. Koethe, L. Hufnagel, and F. A. Hamprecht. Graphical Model for Joint Segmentation and Tracking of Multiple Dividing Cells. *Bioinformatics*, page btu764, Nov. 2014.

[41] M. Schiegg, P. Hanslovsky, B. X. Kausler, and L. Hufnagel. Conservation Tracking. *ICCV 2013*, 2013.

[42] A. Schrijver. *Combinatorial optimization. Polyhedra and efficiency*. Springer, 2003.

[43] J. Shi and J. Malik. Normalized cuts and image segmentation. *TPAMI*, 22(8):888–905, Aug. 2000.

[44] S. Tang, B. Andres, M. Andriluka, and B. Schiele. Subgraph decomposition for multi-target tracking. In *CVPR*, 2015.

[45] R. Tomer, K. Khairy, F. Amat, and P. J. Keller. Quantitative high-speed imaging of entire developing embryos with simultaneous multiview light-sheet microscopy. *Nature Methods*, 9(7):755–763, July 2012.

[46] E. Türetken, C. Becker, P. Glowacki, F. Benmansour, and P. Fua. Detecting irregular curvilinear structures in gray scale and color imagery using multi-directional oriented flux. In *ICCV*, 2013.

[47] E. Türetken, F. Benmansour, B. Andres, H. Pfister, and P. Fua. Reconstructing loopy curvilinear structures using integer programming. In *CVPR*, 2013.

[48] E. Türetken, F. Benmansour, and P. Fua. Automated reconstruction of tree structures using path classifiers and mixed integer programming. In *CVPR*, 2012.

[49] E. Türetken, G. Gonzalez, C. Blum, and P. Fua. Automated reconstruction of dendritic and axonal trees by global optimization with geometric priors. *Neuroinformatics*, 9:279–302, 2011.

[50] X. Wang, E. Turetken, F. Fleuret, and P. Fua. Tracking interacting objects optimally using integer programming. In *ECCV*, 2014.

[51] J. Yarkony and C. Fowlkes. Planar ultrametric rounding for image segmentation. In *NIPS*, 2015.

[52] J. Yarkony, A. Ihler, and C. C. Fowlkes. Fast planar correlation clustering for image segmentation. In *ECCV*, 2012.

## A  Proof of Lemma 2

PROOF If Condition 1 in Def. 2 holds for a set $C \subseteq E$, then all inequalities (1) are satisfied by the $x \in \{0,1\}^E$ such that $x^{-1}(1) = C$. Otherwise, there would exist a $t \in \mathbb{N}$, a cycle $Y$ of $G_t$ and an $e \in Y$ such that $x_e = 1$ and $\forall e' \in Y \setminus \{e\}: x_{e'} = 0$. This implies $|Y \cap C| = 1$, in contradiction to the assumption that $C \cap E_t$ is a multicut of $G_t$. Conversely, if all inequalities (1) are satisfied by an $x \in \{0,1\}^E$, then $C := x^{-1}(1)$ satisfies Condition 1 in Def. 2. Otherwise, there would exist a $t \in \mathbb{N}$ for which $C \cap E_t$ is not a multicut of $G_t$. Thus, there would exist a cycle $Y$ of $G_t$ and an $e \in Y$ such that $Y \cap C = \{e\}$, by definition of a multicut. Hence, the inequality (1) for that cycle $Y$ and that edge $e$ of $Y$ would be violated by $x$. The sufficiency of chordless cycles follows from (1) and is established, *e.g.*, in [16].

If Condition 2 in Def. 2 holds for a set $C \subseteq E$, then all inequalities (2) are satisfied by the $x \in \{0,1\}^E$ such that $x^{-1}(1) = C$. Otherwise, there would exist $t \in \mathbb{N}$, $\{v,w\} \in E_{t,t+1}$ and a path $P \in vw\text{-paths}(G_t^+)$ such that $x_{vw} = 1$ and $x_P = 0$. From $x_{vw} = 1$ follows $\{v,w\} \in E_{t,t+1} \cap C$. From $x_P = 0$ follows that $v$ and $w$ are connected by $P$ in $(V_t^+, E_t^+ \cap \bar{C})$. Both statements together contradict the assumption. Conversely, if all inequalities (2) are satisfied by an $x \in \{0,1\}^E$, then $C := x^{-1}(1)$ satisfies Condition 2 in Def. 2. Otherwise, there would exist $t \in \mathbb{N}$, $\{v,w\} \in E_{t,t+1} \cap C$ and a path $P \in vw\text{-paths}(V_t^+, E_t^+ \cap \bar{C})$. From this follows $x_{vw} = 1$ and $x_P = 0$, in contradiction to the assumption that (2) is satisfied.

If Condition 3 in Def. 2 holds for a set $C \subseteq E$, then all inequalities (3) are satisfied by the $x \in \{0,1\}^E$ such that $x^{-1}(1) = C$. Otherwise, there would exist $t \in \mathbb{N}$, $v_t, w_t \in V_t$, $v_{t+1}, w_{t+1} \in V_{t+1}$, a path $P \in v_{t+1}, w_{t+1}\text{-paths}(G_{t+1})$, and a cut $T \in v_t w_t\text{-cuts}(G_t)$ such that $x_{v_t, v_{t+1}} = 0$ and $x_{w_t, w_{t+1}} = 0$ and $x_P = 0$ and $x_T = 1$. $P$ witnesses the existence of a $v_{t+1} w_{t+1}$-path in $(V, E_{t+1} \cap \bar{C})$. The existence of $T$ certifies the non-existence of a $v_t w_t$-path in $(V, E_t \cap \bar{C})$. Both statements together contradict the assumption. Conversely, if all inequalities (3) are satisfied by an $x \in \{0,1\}^E$, then $C := x^{-1}(1)$ satisfies Condition 3 in Def. 2. Otherwise, there would exist $t \in \mathbb{N}$, $v_t, w_t \in V_t$ and $v_{t+1}, w_{t+1} \in V_{t+1}$ such that $\{v,w\} \in E_{t,t+1} \cap \bar{C}$ and $\{v_{t+1}, w_{t+1}\} \in E_{t,t+1} \cap \bar{C}$ and such that there exist $P \in v_{t+1} w_{t+1}\text{-paths}(V_{t+1}, E_{t+1} \cap \bar{C})$ and $T \in v_t w_t\text{-cuts}(V_t, E_t \cap \bar{C})$. Hence, $x_{v_t, v_{t+1}} = 0$ and $x_{w_t, w_{t+1}} = 0$ and $x_P = 0$ and $x_T = 1$, in contradiction to the assumption that (3) is satisfied. □